\pdfoutput=1

\documentclass[11pt]{article}

\usepackage{acl}
\usepackage{amssymb}
\usepackage{times}
\usepackage{latexsym}

\usepackage[T1]{fontenc}

\usepackage[utf8]{inputenc}

\usepackage{microtype}

\usepackage{multirow}
\usepackage{booktabs}
\usepackage{arydshln}
\usepackage{caption}
\usepackage{graphicx}
\usepackage{amsmath}
\DeclareMathOperator*{\argmax}{arg\,max}

\usepackage{algorithm,algpseudocode}

\let\emptyset\varnothing
\title{Explicit Syntactic Guidance for Neural Text Generation}

\author{ 
 Yafu Li$^{\spadesuit \clubsuit}$\footnotemark[1]\hspace{0.5mm}, 
 Leyang Cui$^{\heartsuit}$\footnotemark[2]\hspace{0.5mm}, 
 Jianhao Yan$^{\spadesuit \clubsuit}$\hspace{0.5mm}, Yongjing Yin$^{\spadesuit \clubsuit}$\hspace{0.5mm} \\
  \bf{
 Wei Bi$^{\heartsuit}$\hspace{0.5mm},  
 Shuming Shi$^{\heartsuit}$\hspace{0.5mm}, 
 Yue Zhang$^{\clubsuit \diamondsuit}$}\footnotemark[2]\hspace{0.2mm}\hspace{1.5mm} \\
 $^\spadesuit$ Zhejiang University \ \ \ $^\heartsuit$ Tencent AI lab \\
 \quad$^\clubsuit$ School of Engineering, Westlake University\\
 \quad$^\diamondsuit$ Institute of Advanced Technology, Westlake Institute for Advanced Study\\ 
 \texttt{yafuly@gmail.com} \\
 \quad\texttt{\{leyangcui,victoriabi,shumingshi\}@tencent.com} \\
 \quad\texttt{\{yanjianhao,yinyongjing,zhangyue\}@westlake.edu.cn}\\
}
\date{}
\begin{document}
\maketitle
{
\renewcommand{\thefootnote}{\fnsymbol{footnote}}
\footnotetext[1]{\ Work was done during the internship at Tencent AI lab.}
\footnotetext[2]{\ Corresponding authors.}
}
\begin{abstract}
Most existing text generation models follow the sequence-to-sequence paradigm. 
\textit{Generative Grammar} suggests that humans generate natural language texts by learning language grammar.
We propose a syntax-guided generation schema, which generates the sequence guided by a constituency parse tree in a top-down direction. 
The decoding process can be decomposed into two parts: 
(1) predicting the infilling texts for each constituent in the lexicalized syntax context given the source sentence; 
(2) mapping and expanding each constituent to construct the next-level syntax context.
Accordingly, we propose a structural beam search method to find possible syntax structures hierarchically.
Experiments on paraphrase generation and machine translation show that the proposed method outperforms autoregressive baselines, 
while also demonstrating effectiveness in terms of interpretability, controllability, and diversity.

\end{abstract}

\section{Introduction}


Natural language generation (NLG), 
such as paraphrase generation \cite{sun-etal-2021-aesop},
text summarization \cite{text-sum}, machine translation \cite{vaswani2017attention, Edunov:emnlp18}, and language models~\cite{gpt3, gpt4},
have shown remarkable progress in the past few years. 
Most of the highest-performing NLG models train the model based on source-target correspondence and conduct autoregressive inference,
which achieves competitive empirical performances yet deviates from a range of desirable attributes of human language generation, e.g., lack of interpretability~\cite{alvarez-melis-jaakkola-2017-causal, he-etal-2019-towards,  aaai-LiY21}.
%

It has been shown that humans generate language by learning and manipulating language grammar \cite{meantext,montague1974d},
which generative grammar \cite{chomsky1965aspects} considers as a finite rule set that combines words to form grammatical sentences, thereby avoiding enumeration of surface sequences, which can significantly increase data sparsity and reduce learning efficiency.
In this process, syntax plays a crucial role, imposing constraints on how to construct sentences.
Syntax knowledge has been found \textit{implicitly} contained by deep neural models~\cite{bert-syntax1, clark2019what} and also useful for NLG tasks~\cite{yang2020improving, sun-etal-2021-aesop, xie2021transformer}. 
However, relatively little recent work has considered \textit{explict} syntax in NLG \cite{con-tree-decoder-nmt}.

\begin{figure}[t]
\centering
\includegraphics[width=1.0\linewidth]{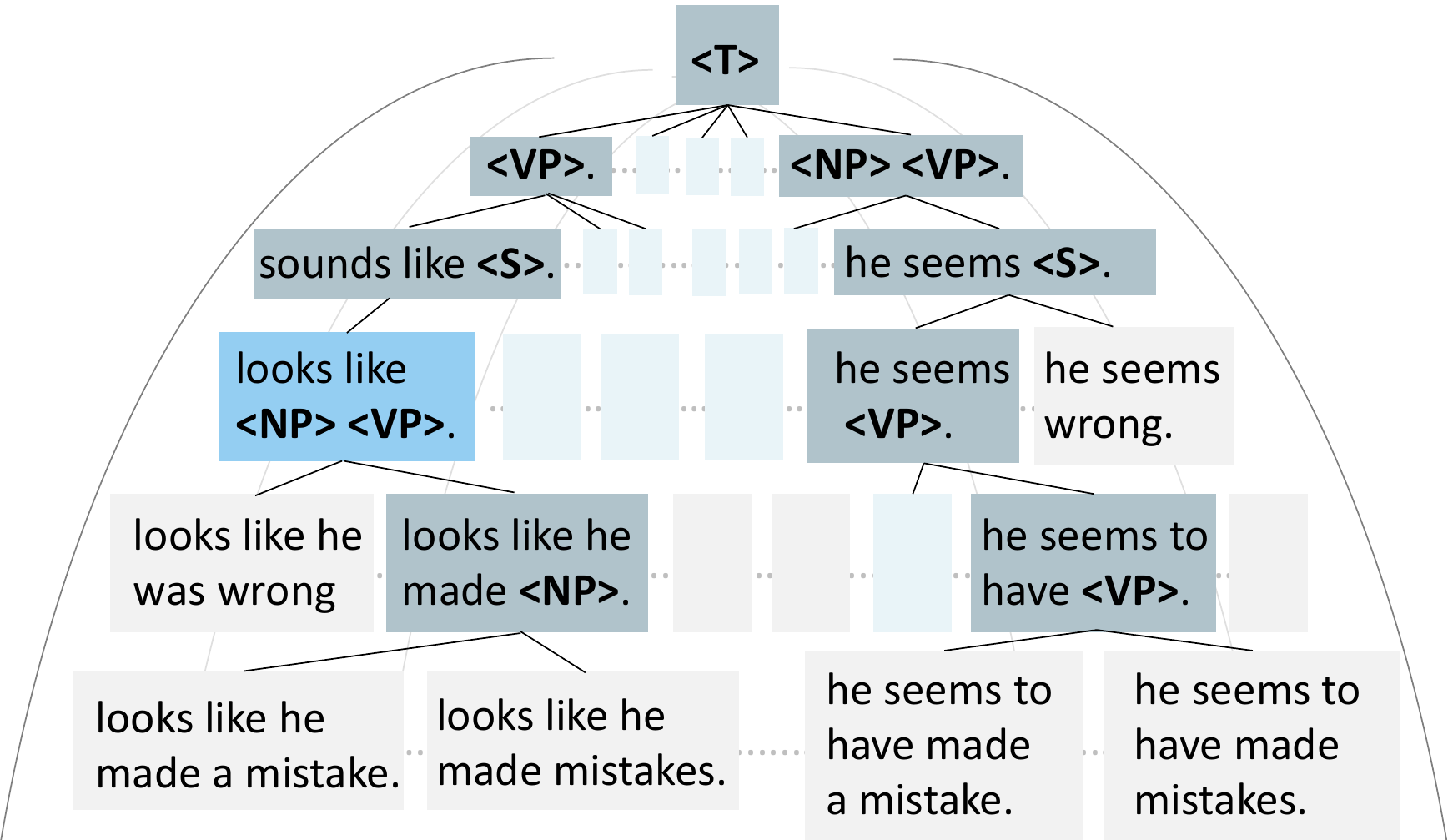}
    \caption{Syntax-guided generation: searching the hypotheses hierarchically throughout the syntax tree in a top-down direction, starting from the root node ``<\texttt{T}>''. The green blocks denote the possible syntax structures at different tree depths, the blue one denotes the external modification, whereas the gray ones denote the finalized hypotheses, marking the end of search paths.}
    \label{fig:intro}
\end{figure}

Inspired by the above psycholinguistic observation, 
we propose a syntax-guided generation scheme, which generates text by following a well-defined grammar.
As shown in Figure~\ref{fig:intro}, instead of sequential generation, the model generates the sentence in a hierarchically top-down manner guided by the constituency parse tree, starting with the root node <\texttt{T}>.
Syntactic categories such as noun phrases <\texttt{NP}> and verb phrases <\texttt{VP}> are integrated with tokens in the generation process, and 
the model simultaneously considers multiple syntax structures at each tree depth, hierarchically exploring the syntax tree for reasonable hypotheses.

Intuitively, such a generation paradigm has the following advantages compared with autoregressive generation.
First,
akin to the language learning process of human beings, grammar learning breaks down non-enumerable surface sequences into finite pieces, acting as a training curriculum.
Second,
it provides an effective and interpretable pathway to probe into the generation process.
Consequently, generation errors can be traced back to specific constituent expansion at the respective tree depth.
Third,
one can manipulate the generation process by exerting versatile control at arbitrary depths, 
e.g., modifying the translation of a verb phrase and constraining the paraphrase style with syntax templates.
Forth,
diverse sequences can be generated by exploring various syntax structures hierarchically throughout the syntax tree.

We implement the above process on Transformer \cite{vaswani2017attention}.
As shown in Figure~\ref{fig:intro}, 
the generation process proceeds under the guidance of syntactic grammar.
Starting from the root node ``<\texttt{T}>'', 
the model recursively generates the infilling texts (e.g., ``he'' and ''seems <\texttt{S}>'') for each constituent in the current lexicalized syntax context (e.g, ``<\texttt{NP}> <\texttt{VP}>.''.), 
and infills each one accordingly to construct the next-level lexicalized syntax context (e.g., ``he seems <\texttt{S}>.'').
The generation proceeds until there is no remaining constituent.
The infilling texts are predicted by a Transformer-based model, which is trained by maximizing the likelihood of infilling texts for each constituent in the syntax context based on the source input.
To explore more syntactically diverse and reasonable hypotheses during inference, we propose \textit{structural beam search},
which searches promising syntax structures over the entire syntax tree in a top-down manner, as shown in Figure~\ref{fig:intro}.

To isolate the effect of syntax and avoid the influence of other transformation factors, 
we conduct experiments on two sequence-to-sequence (seq2seq) tasks \textit{with semantic equivalence} between the source and target sequences: paraphrase generation and machine translation. 
Empirical results demonstrate that our method can generate sequences with higher quality than the seq2seq baselines.
Quantitative analysis demonstrates that the generation process can be interpreted effectively.
In addition, our method demonstrates the capability of executing control from both syntax templates and fine-grained manual modifications.
Finally, we show the diversity advantage through both automatic evaluation and human evaluation.
We release the code on \url{https://github.com/yafuly/SyntacticGen}.

\section{Related Work}
\paragraph{Syntax as Extra Input.}
A line of work incorporates syntax knowledge as extra input to boost task performance.
In paraphrase generation,
\citet{iyyer-etal-2018-adversarial}, \citet{chen-etal-2019-controllable}, \citet{kumar-etal-2020-syntax} and \citet{sun-etal-2021-aesop} additionally encode a constituency tree to produce controllable paraphrases.
For machine translation, 
researchers utilize syntactic information to boost the neural machine translation system using syntactic encoders \cite{syntax-encoder,ma2018forest,eriguchi2019incorporating,ma2020syntax,yang2020improving}, position encoding~\cite{ma2019improving,xie2021transformer}, attention mechanism~\cite{chen2018syntax,peng2019neural}, and auxiliary training objectives \cite{ma2019improving}.

\paragraph{Syntax for Generation Guidance.} 
Different from the above work, we focus on guiding generation explicitly following syntactic grammar.
Typically, 
\citet{aharoni-goldberg-2017-towards} and \citet{le2017improving} learn the mapping from sequences to linearized constituency trees to improve machine translation.
\citet{eriguchi2017learning} proposes a hybrid decoder with RNNG~\citep{dyer-etal-2016-recurrent} to jointly learn parse actions and word predictions.
\citet{dep-tree-decoder-nmt} and \citet{con-tree-decoder-nmt} design a syntactic tree decoder based on LSTM \cite{lstm}, with an extra rule decoder. 
\citet{yang2020improving1} introduce a syntax-guided soft target template as extra prompts in Transformer.
Different from their work, our method leverages Transformer strengths and breaks down the sequence-to-sequence generation process into a hierarchically top-down generation guided by the syntax tree.

\begin{figure*}[t!]
\centering
\includegraphics[width=0.99\linewidth]{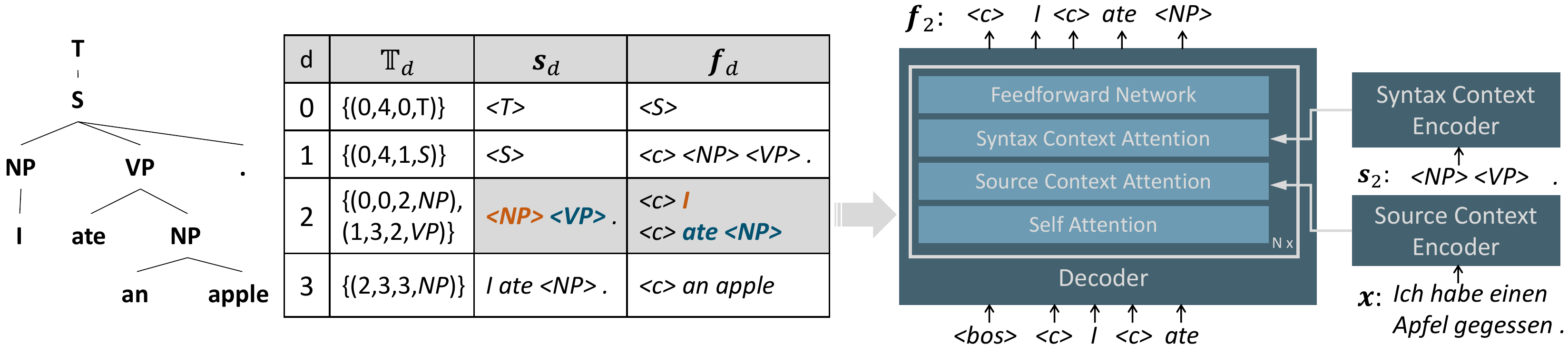}
\caption{
\label{fig:method}
Method illustration: the left part demonstrates the construction of the training triplet (i.e., $(\mathbf{x},\boldsymbol{s}_d,\boldsymbol{f}_d)$) based on the constituency parse tree; the right part denotes the architecture of the neural decoder, which takes in the German source sentence $\mathbf{x}$ and the syntax context $\boldsymbol{s}_2$ as input, and predicts the infilling text $\boldsymbol{f}_2$.}
\end{figure*}

\section{Method}

\subsection{Baseline Transformer}
Transformer models the correspondence between the source sequence $\mathbf{x}=\{x_1,\dots,x_{|\mathbf{x}|}\}$ and the target sequence $\mathbf{y}=\{y_1,\dots,y_{|\mathbf{y}|}\}$ in an end-to-end fashion. 
The Transformer encoder transforms the discrete source sequence $\mathbf{x}$ into a continuous representation, which the Transformer decoder utilizes to generate the target sequence.
The conditional probability $p(\mathbf{y}|\mathbf{x})$ can be factorized in an autoregressive way:
\begin{equation}
   p_{\boldsymbol{\theta}}(\mathbf{y}|\mathbf{x}) = \prod_{t=1}^{|\mathbf{y}|} p_{\boldsymbol{\theta}}(y_t|\mathbf{x},y_{1:t-1}),
\end{equation}
where $\boldsymbol{\theta}$ denotes the model parameters.

Given a source-target training set $\mathcal{D}=\{\mathbf{x}^i, \mathbf{y}^i\}|_{i=1}^{|\mathcal{D}|}$, the model is optimized by minimizing the cross-entropy (CE) loss:
\begin{equation}
    \mathcal{L}_{ce}^{\mathcal{D}} = - \sum_{i=1}^{|\mathcal{D}|} \sum_{t=1}^{T} \log p_{\boldsymbol{\theta}}(y_t^i|\mathbf{x}^i,y_{1:t-1}^i).
\label{eq:dialogue_loss}
\end{equation}

\subsection{Syntax-guided Generation}
In this section, 
we introduce syntax-guided generation,
which generates texts by hierarchically expanding constituents in syntax contexts throughout the syntax tree,
while also leveraging the strengths of Transformer.
In general, 
the generation process can be decomposed into two stages: 
(1) \textbf{neural generation}: the neural decoder (Section \ref{sec:neu_gen}) generates the infilling sequences based on the source sequence and the syntax context; 
(2) \textbf{constituent expansion}: predicted infilling sequences are mapped and filled into each constituent in the syntax context accordingly (Section \ref{sec:gen_process}), forming the next-level syntax context.
To facilitate parallelism during training, we decompose the sequence-to-sequence dataset to a triplet set, where the neural decoder is optimized to maximize the probability of the infilled sequence (e.g., "<\texttt{c}> I <\texttt{c}> ate <\texttt{NP}> .") given the lexicalized syntax context (e.g., "<\texttt{NP}> <\texttt{VP}> ."), as shown in Figure \ref{fig:method}.

\subsubsection{Triplet Construction}
Given a target sequence $\mathbf{y}$, the corresponding constituency parse tree of depth $|\mathbb{T}|$ can be composed by a set of labeled spans $\mathbb{T}$:
\begin{equation}
\small
    \mathbb{T} = \{\mathbb{T}_d\}|_{d=1}^{|\mathbb{T}|} = \{\{(a_k,b_k,d,l_k)\}|_{k=1}^{|\mathbb{T}_d|}\}|_{d=1}^{|\mathbb{T}|},
\end{equation}
where $a_k$ and $b_k$ represent the $k$-{th} constituent span's fencepost positions at depth $d$, and $l_k$ represents the constituent label. 
Our model is optimized to predict the next-level span sets $\mathbb{T}_{d}$ given the previous one and the source input,
i.e., $p_{\boldsymbol{\theta}}(\mathbb{T}_{d}|\mathbb{T}_{d-1},\mathbf{x})$.

Given the set of labeled spans at depth $d$, i.e., $\mathbb{T}_d$, we transform the target sequence into a lexicalized syntax sequence of length $|\boldsymbol{s}_d|$: $\boldsymbol{s}_d=\{s_{d;1},s_{d;2},\dots,s_{d;|\boldsymbol{s}_d|}\}$, by keeping the lexical tokens and replacing the constituent spans with corresponding labels.
For instance, the sequence ``I ate an apple .'' is transformed to $\boldsymbol{s}_2=$\{<\texttt{NP}>,<\texttt{VP}>,.\} at depth 2, and is transformed to $\boldsymbol{s}_3=$\{I,ate,<\texttt{NP}>,.\} at depth 3, as shown in Figure \ref{fig:method}. 
The alignment between $\boldsymbol{s}_2$ and $\boldsymbol{s}_3$ can be modeled as a text-infilling task. 
For example, the \{<\texttt{NP}>\}, \{<\texttt{VP}>\} and at depth 2 are replaced by \{I\} and \{ate <\texttt{NP}>\} at depth 3, respectively.
To generate the whole $\boldsymbol{s}_3$ based on $\boldsymbol{s}_2$ in one pass, we concatenate all the infilling texts with a special token ``<\texttt{c}>'', yielding an infilling sequence $\boldsymbol{f}_2=$ \{<\texttt{c}>,I,<\texttt{c}>,ate,<\texttt{NP}>\}.

Similarly for each syntax context $\boldsymbol{s}_d$,
we collect the respective infilling texts for each constituent in the lexicalized sequence at depth $d+1$, and concatenate them to construct the target infilling sequence of length $|\boldsymbol{f}_d|$: $\boldsymbol{f}_d = \{f_{d;1},f_{d;2},\dots,f_{d;|\boldsymbol{f}_d|}\}$. 
In this way, a triplet is constructed for a source-target sequence pair at depth $d$: $\{(\mathbf{x},\boldsymbol{s}_d,\boldsymbol{f}_d)\}$.
We traverse the target syntax tree in level-order to obtain the full set $\Phi$ of training triplets for a training instance:
\begin{equation}
    \Phi =  \{\Phi_d\}|_{d=1}^{|\mathbb{T}|-1} = \{(\mathbf{x},\boldsymbol{s}_d,\boldsymbol{f}_d)\}|_{d=1}^{|\mathbb{T}|-1}.
\end{equation}

Given a sequence-to-sequence training set $\mathcal{D}=\{\mathbf{x}^i, \mathbf{y}^i\}|_{i=1}^{|\mathcal{D}|}$, we go through the full training set to construct the complete triplet set $\Psi$:
\begin{equation}
    \Psi = \{\Phi^i\}|_{i=1}^{|\mathcal{D}|} = \{(\mathbf{x}^j,\boldsymbol{s}^j,\boldsymbol{f}^j)\}|_{j=1}^{\sum _{i=1}^{|\mathcal{D}|}|\Phi^i|}.
\end{equation}

\subsubsection{Neural Decoder}
\label{sec:neu_gen}
Given a triplet instance $\Psi^j$, we construct the \textbf{neural decoder} based on Transformer to model the generative probability $p_{\boldsymbol{\theta}}(\boldsymbol{f}^j|\mathbf{x}^j,\boldsymbol{s}^j)$. 
The neural decoder takes the source sequence and the lexicalized syntax context as input and generates the corresponding infilling texts, as shown in Figure \ref{fig:method}.

Besides the encoder that encodes source context, we introduce an extra Transformer encoder, i.e., syntax context encoder, to encode the lexicalized syntax context into a representation.
On top of self-attention and source context attention, we insert an extra attention layer (syntax context attention) into each decoder layer to incorporate syntax contexts, as shown in the right part of Figure~\ref{fig:method}.


Similarly, the probability of the infilling sequence can be factorized as:
\begin{equation}
    p_{\boldsymbol{\theta}}({\boldsymbol{f}|\mathbf{x},\boldsymbol{s}})=\prod_{t=1}^{|\boldsymbol{f}|}p_{\boldsymbol{\theta}}(f_{t}|\mathbf{x},\boldsymbol{s},f_{1:t-1}).
\label{eq:seq_prob}
\end{equation}

We define the scoring function for an infilling sequence as the sum of the log probabilities:
\begin{equation}
    \mathrm{score}(\mathbf{x},\boldsymbol{s},\boldsymbol{f})=\sum_{t=1}^{|\boldsymbol{f}|}\log p_{\boldsymbol{\theta}}(f_t|\mathbf{x},\boldsymbol{s},f_{1:t-1}).
\label{eq:score}
\end{equation}

We adopt the standard cross-entropy loss (CE loss) to optimize our model, where the loss for the $j$-th triplet in the training set $\Psi$ can be written as:
\begin{equation}
    \mathcal{L}_{ce}^{j} = -  \sum_{t=1}^{|\boldsymbol{f}^{j}|}  \log p_{\boldsymbol{\theta}}(f_{t}^j|\mathbf{x}^{j},\boldsymbol{s}^{j},f_{1:t-1}^{j}),
\end{equation}
and the CE loss across the whole triple set $\Psi$ becomes:
\begin{equation}
    \mathcal{L}_{ce}^{\Psi} = \sum_{j=1}^{|\Psi|} \mathcal{L}_{ce}^{j}.
\end{equation}

\subsubsection{Generation Process}
\label{sec:gen_process}
Given a source sequence, our model generates the target sequence in a top-down manner which is grounded on syntactic grammar rules.
As shown in Figure \ref{fig:method}, the neural decoder first encodes the source sequence $\mathbf{x}$ into the source context representation $\mathbf{h}_{src}$, which remains fixed and can be reused throughout the generation process.
Initially, the neural decoder generates the infilling sequences $\mathbf{t}_0$ given $\boldsymbol{x}$ and $\boldsymbol{s}_0=$\{<\texttt{T}>\}, based on Equation \ref{eq:seq_prob}.
Then the model proceeds with the generation process via iteratively generating infilling texts and expanding constituents.

At each iteration step (i.e., tree depth),
the neural decoder generates the infilling sequence $\boldsymbol{f}_d$ for the syntax context $\boldsymbol{s}_d$:
\begin{equation}
    \boldsymbol{f}_d = \argmax_{\boldsymbol{f'}}p_{\boldsymbol{\theta}}({\boldsymbol{f'}|\mathbf{x},\boldsymbol{s}_d})
\end{equation}
Then the constituent expansion function yields the next-level syntax context given the syntax context and the infilling sequences predicted by the neural decoder:
\begin{equation}
\boldsymbol{s}_{d+1}=\mathrm{expand}(\boldsymbol{s}_d,\boldsymbol{f}_d).
\end{equation}

Specifically, we first separate the infilling sequences by the special separator ``<\texttt{c}>'' into a group of infilling texts, e.g., 
spliting $\boldsymbol{f}_2=$\{\{<\texttt{c}>,I,<\texttt{c}>,ate,<\texttt{NP}>\}\} to \{\{I\},\{ate <\texttt{NP}>\}\}.
Then we fill in each of the infilling texts into the corresponding constituent in the syntax context $\boldsymbol{s}_2$ to obtain the syntax context at the following level, e.g., $\boldsymbol{s}_3$=\{I,ate,<\texttt{NP}>,.\}.
The syntax context encoder encodes the updated syntax context $\boldsymbol{s}_{d+1}$ and starts the next iteration.
The remaining decoding process loops between these two stages, 
until there is no constituent label in the syntax context, or a maximum tree depth is reached, as shown in Figure \ref{fig:method}.

As the model behavior on expanding constituents over the entire syntax tree is completely accessible, 
the generation process can be effectively interpreted, as shown in Section \ref{sec:ana_inter}.
Moreover, manual modifications can be directly incorporated into the expansion process for each constituent throughout the syntax tree (Section \ref{sec:ana_control}).
Finally, more than one syntax structure can be considered simultaneously at each tree depth, enabling searching for hypotheses of better syntactical diversity(Section \ref{sec:ana_div}).


\begin{figure*}[t]
\small
\centering
\includegraphics[width=1\linewidth]{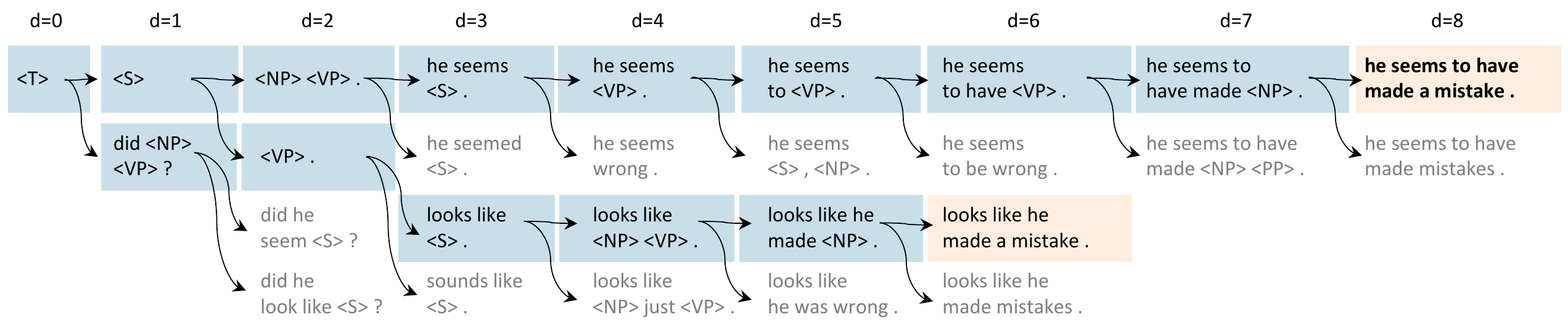}
    \caption{A real example of our model generating a paraphrase given the source sequence ``it seems like he has made a mistake.'', under the structural beam search of width 2. 
    Diverse syntax structures are explored during the generation, e..g, ``<\texttt{VP}>.'', ``<\texttt{NP}> <\texttt{VP}>.'', and ``did <\texttt{NP}> <\texttt{VP}>?''.
    }
    \label{fig:beam}
\end{figure*}

\subsubsection{Structural Beam Search}
By default, our model selects the best infilling texts greedily in each iteration.
We introduce \textbf{structural beam search} to explore the hypothesis space for a more accurate and diverse generation.
Similar to standard beam search~\cite{seq2seq}, structural beam search maintains a beam width of candidates at each iteration. 
Thanks to explicitly traversing the constituency parse tree during inference, our method is able to search promising syntax structures throughout the syntax tree in a top-down manner.
We show a real example of our model generating a paraphrase in Figure \ref{fig:beam}.

At each level, we apply standard beam search for neural generation and keep top $k$ infilling texts along with their scores, computed by Equation \ref{eq:score}.
Taking previous predictions into consideration, we introduce a moving average mechanism to trade off confidence between the predictions from lower levels and the current-level prediction. 
Specifically, suppose $\boldsymbol{s}_{i}$ is the $i$-th syntax context in the $k$-width beam at the current depth, with an accumulated score of $\delta_{\boldsymbol{s}_i}$; and $\boldsymbol{f}_{j;\boldsymbol{s}_i}$ is the $j$-th infilling sequence candidate from the neural generation beam given the syntax context $\boldsymbol{s}_i$, with a score of $\delta_{\boldsymbol{f}_{j;\boldsymbol{s}_i}}$.
A beam of next-level syntax contexts is constructed, by filling in the current syntax context with the corresponding infilling sequences:
\begin{equation}
    \boldsymbol{s}_{ik+j}=\mathrm{expand}(\boldsymbol{s}_i,\boldsymbol{f}_{j;\boldsymbol{s}_i}).
    \label{eq:updata_text}
\end{equation}

The updated score for 
each of the next-level syntax contexts in the beam is given by:
\begin{equation}
    \delta_{ik+j} = \alpha \delta_{\boldsymbol{s}_i} + (1-\alpha) \delta_{\boldsymbol{f}_{j;\boldsymbol{s}_i}},
    \label{eq:weight_beam}
\end{equation}
where $\alpha$ is a hyper-parameter (\textbf{accumulation weight}) that determines how much weight is put on predictions at lower levels.
Then the beam is further pruned by their updated scores to maintain the beam width.
For example, the first two candidate syntax contexts are selected at depth 2 in Figure \ref{fig:beam}.
Algorithm implementation details can be referred to in Appendix \ref{app:alg}.

\begin{table*}[t!]
\small
    \scriptsize{\centering
        \begin{tabular}{cccccccccc}
            \toprule
              \textbf{Model} & \textbf{BLEU}$\uparrow$  / \textbf{self-BLEU}$\downarrow$  / \textbf{iBLEU}$\uparrow$ & \textbf{METEOR}$\uparrow$ & \textbf{ROUGE-1/2/L}$\uparrow$  & $\mathbf{D}_{lex}$$\uparrow$  &  $\mathbf{D}_{syn}\uparrow$ \\
            \midrule
                Copy & 18.5 / 100 / -17.1 & 28.8 & 50.6 / 23.2 / 47.7 & 0.0 & 0.0 \\
                Gold  & 100.0 / 18.6 / 64.4 & 100.0 &100.0 /100.0  / 100.0  & 20.7 &  32.6\\
            \midrule                
              \multicolumn{10}{c}{\textbf{without Syntax Control}}\\
             
            \midrule 
SCPN \cite{iyyer-etal-2018-adversarial} & 12.1 / - / - & 23.3 & 35.7 / 15.1  / 32.9 & - & - \\ 
                AESOP \cite{sun-etal-2021-aesop}  & 15.0 / - / - & 26.1 & 47.0 / 21.3 / 47.3  & - & - \\
            \hdashline
                Transformer (beam 1) & 15.2 / 28.2 / 2.2 & 29.5 & 49.8 / 23.6 / 49.2 & 17.4 &  19.8\\
                Our Method (beam 1) & 18.6 / \textbf{15.2} / 8.5 & 30.8 & 51.1 / 26.3 / 51.3 & \textbf{21.6} & 24.4 \\
                Transformer (beam 5) & 17.6 / 33.8 / 2.2 & 31.1 & \textbf{51.9} / 26.0 / 51.0 & 16.2 & 18.1\\
                Our Method (beam 5) & \textbf{19.3} / 16.4 / \textbf{8.6} & \textbf{31.5} & 51.8 / \textbf{27.0} / \textbf{52.2} & 21.5 & \textbf{25.1}  \\
            \midrule  
              \multicolumn{10}{c}{\textbf{with Human-annotated Syntax Control}} \\
            \midrule  
             CGEN \cite{chen-etal-2019-controllable}  & 13.6 /- /- & 24.8 & 44.8 / 21.0 / 48.3  & - & - \\
            SGCP-F \cite{kumar-etal-2020-syntax} &  15.3 / - / -& 25.9 & 46.6 / 21.8 / 49.7  & - & - \\
            SGCP-R \cite{kumar-etal-2020-syntax} & 16.4 / - / - & 28.8& 49.4 / 22.9 / 50.3  & - & - \\
              AESOP-F \cite{sun-etal-2021-aesop} & 20.4 / - /- & 30.0 & 52.0 / 27.8 / \textbf{55.3}   & - & -\\
              \hdashline
               Our Method & \textbf{20.9} / \textbf{10.5} /  \textbf{13.0} & \textbf{33.3} & \textbf{54.1} / \textbf{29.7} / \textbf{55.3} & \textbf{22.6} & \textbf{27.7}  \\

            \bottomrule
        \end{tabular}
        \caption{Experimental results on paraphrase generation (ParaNMT-small).}
        \label{tab:para}
    }
\end{table*}

\begin{table*}[t!]
\small
\scriptsize{
    \centering
    \begin{tabular}{ccccccccccc}
    \toprule
         \multirow{2}{*}{\bf Model} &  \multicolumn{6}{c}{\textbf{NIST Zh-En}} & \multirow{1}{*}{\textbf{WMT16}} & \multicolumn{2}{c}{\textbf{WMT14}}  \\
         & MT02 & MT03 &  MT04 & MT05 & MT08 & avg & Ro-En  & En-De & De-En  \\
    \midrule
         Transformer (beam 1) & 48.9 & 49.2 & 50.7 & 49.3 & 41.4 & 47.9 & 33.9 & 27.9  & 30.7 \\ 
         Our Method (beam 1) & 50.8 & 51.8 & 51.9 & 51.7 & 42.2 & 49.7 & 34.4 & 28.6 & 31.8\\
         Transformer (beam 5) & 49.8 & 50.1 & 51.1 & 50.1 & 42.3 & 48.7 & 34.1 & 28.3 & 31.3 \\
         Our Method (beam 5) &\textbf{51.1} &\textbf{52.4} & \textbf{52.4} &\textbf{52.1} & \textbf{43.1} & \textbf{50.2} &\textbf{34.9} &  \textbf{28.7} &  \textbf{32.2}\\
     \bottomrule
    \end{tabular}
    \caption{Experimental results (BLEU score) on machine translation benchmark datasets. The result of our method is statistically significant compared to the corresponding Transformer baseline with $p<0.05$ \cite{sig_test}.} 
    \label{tab:mt}
}
\end{table*}

\section{Experiment Setup}
\paragraph{Datasets}
For paraphrase generation, we experiment on ParaNMT-small \cite{chen-etal-2019-controllable}, which contains 500K sentence-paraphrase pairs for training, 500 for validation, and 800 for testing. Both validation and test sets are provided with human-annotated sentence exemplars from which syntax information can be extracted for controlling paraphrase generation.
For machine translation, we use NIST Chinese-English (Zh-En), WMT'16 Romanian-English (Ro-En), WMT'14 English-German (De-En), and WMT'14 English-German (En-De). 
For WMT datasets, we follow the official split for validation and testing.
For NIST Zh-En, we use MT06 as the validation set and choose MT02, MT03, MT04, MT05, and MT08 as the test sets.
For all datasets, we use Berkeley Parser \cite{kitaev-klein-2018-constituency,kitaev-etal-2019-multilingual} to obtain constituency parse trees and use the most frequent constituents (e.g., <\texttt{NP}>, <\texttt{VP}>, <\texttt{PP}> and <\texttt{S}>) for syntactic guidance.
\paragraph{Model Settings}
For Transformer baselines, we adopt the Transformer\_Base configuration which consists of a 6-layer encoder and decoder.
For our model, 
we keep the 6-layer source context encoder, and set the number of layers for both the syntax context encoder and the decoder as 3, resulting in a similar model size with Transformer\_Base.
The accumulation weight $\alpha$ is as 0.8 for structural beam search based on validation experiments.
For machine translation, we adopt sequence-level distillation \cite{seq_kd} for both our model and the corresponding baseline Transformer.
More details are shown in Appendix \ref{app:setup}.

\paragraph{Evaluation} We use the BLEU score \cite{bleu} to evaluate machine translation performance. For paraphrase generation, we also adopt ROUGE \cite{rouge} and METEOR \cite{meteor} as reference-based metrics.
Besides, we report iBLEU \cite{ibleu}:
\begin{align*}
    \mathrm{iBLEU} = r \cdot \mathrm{BLEU}(\mathrm{hypothesis},\mathrm{reference}) \\
    - (1-r)\cdot \mathrm{BLEU}(\mathrm{hypothesis},\mathrm{source}),
\end{align*}
which evaluates the generation fidelity with novelty to the source sentence considered\footnote{r is set as 0.7.}.
Following \citet{quality_acl_2022},
we consider two reference-free metrics: (1) lexical diversity score, i.e., $\mathbf{D}_{lex}$, which is the normalized character-level minima edit distance between the bag-of-words; and (2) syntax diversity score, i.e., $\mathbf{D}_{syn}$, which is the normalized tree edit distance.
Both scores measure generated paraphrases with the source sequences unless specified.

\section{Results}

\begin{table}[t]
\centering
\small
\begin{tabular}{cccc}
    \toprule
     \textbf{Model} & \textbf{iBLEU}$\uparrow$& $\mathbf{D}_{lex}$$\uparrow$ & $\mathbf{D}_{syn}$$\uparrow$  \\
     \midrule
     BART &  4.4  & 19.6    &  24.4  \\
     BART + Our Method &  \textbf{8.8} &  \textbf{21.3}   &  \textbf{24.7}  \\
\bottomrule
\end{tabular}
    \caption{Results for training on BART, compared with sequence-to-sequence BART for paraphrase generation.}
    \label{tab:bart}
\end{table}
\paragraph{Paraphrase}
We compare our method with the baselines and previous work on syntax-control paraphrase generation.
Another two baselines are also listed, i.e., copy the source input and use the reference as the output. 
The results are shown in Table \ref{tab:para}.
For paraphrase generation \textbf{without syntax control} (the center section in Table \ref{tab:para}), our method achieves higher performance than the seq2seq Transformer, in both greedy and beam search settings.
Typically, our method under greedy decoding obtains comparable results with the Transformer under beam search, and even outperforms under some metrics.
The advantage of our method becomes larger for metrics such as iBLEU, $\mathbf{D}_{lex}$, and $\mathbf{D}_{syn}$, which consider generation novelty compared with the source input.
For example, compared with Transformer (beam 5), our method (beam 5) gives a much lower self-BLEU score (\textbf{16.4} v.s. \textbf{33.8}) and higher diversity scores (\textbf{21.5} v.s. \textbf{16.2} for lexical diversity and \textbf{25.1} v.s. \textbf{18.1} for syntax diversity), indicating better generation diversity and contributing to a significant improvement on iBLEU (\textbf{8.6} v.s. \textbf{2.2}).
\textbf{With annotated exemplars} (the lower section in Table \ref{tab:para}), our model obtains further improvement over the non-exemplar setting and achieves better performance compared to previous work which utilizes full syntactic parse.

We extend our method to the \textbf{pre-trained language model} (PLM) setting and present the result in Table \ref{tab:bart} (Details in Appendix \ref{app:bart}). It can be seen from the table that the utilization of BART~\cite{bart} improves the generation diversity for the sequence-to-sequence model significantly.
Despite the narrowed gap, our model outperforms the seq2seq counterpart in terms of iBLEU and lexical diversity by a considerable margin.

 \paragraph{Machine Translation}
As shown in Table \ref{tab:mt}, our method achieves consistent performance (BLEU score) improvement over the Transformer baseline.
The improvement is larger for the greedy setting (+1.5 BLEU scores on average), compared with the beam search setting (+1.2).
This indicates that using syntax to guide and constrain generation yields more reasonable and high-quality hypotheses than the greedy autoregressive generation, and thus relies less on search algorithms (e.g., beam search).
Note that compared with the English-oriented datasets, our model obtains a smaller performance improvement on WMT'14 En-De.
This can be because the German parser is less accurate than the English one (92.1 v.s. 96.3 for F1 score), resulting in a training set with lower quality.

\section{Analysis}
We first discuss the influence of grammar quality, 
then we understand the potential advantages of our method from three perspectives, i.e., interpretability, controllability, and diversity.

\subsection{The Influence of Grammar Quality}
\label{sec:grammar}
Intuitively, learning syntactic grammar of higher quality results in better generation performance, e.g., the advantage of our method on English-oriented datasets is larger than the German-oriented one.
To further explore the influence of grammar quality,  
we randomly replace a certain ratio of the constituent labels with a random one to simulate a less accurate parser.
We conduct experiments on the WMT'16 Ro-En dataset.
By injecting noise of ratios of \textbf{0.2} and \textbf{0.4}, the model performance deteriorates from 34.9 to \textbf{34.6} and \textbf{32.3} accordingly, indicating  the quality of syntactic grammar exerts a large influence on model's generation performance.

\begin{table}[t]
\centering
\small
\begin{tabular}{cccc}
    \toprule
     \textbf{Dataset} & \textbf{Precision}& \textbf{Recall} &\textbf{ F1 Score}  \\
     \midrule
     ParaNMT-small & 96.0\% & 98.4\%   &  97.2\%  \\
     NIST Zh-En &  96.6\% &  96.8\%   &  96.7\%  \\
    WMT'16 Ro-En &  93.5\% & 94.2\%   & 93.9\%  \\
    WMT'14 De-En &  95.7\% &  96.3\%  &  96.0\% \\
    WMT'14 En-De &  84.4\% &  95.4\%  &  89.6\%  \\
\bottomrule
\end{tabular}
    \caption{The quantitative evaluation of the models' interpretability.}
    \label{tab:results_parser}
\end{table}

\subsection{Interpretability}
\label{sec:ana_inter}

We evaluate the model's interpretability based on its capability of providing explanations in understandable terms to a human \cite{DoshiVelez2017TowardsAR},
i.e., 
whether it generates texts following language grammar.
We trace each constituent expansion during generation and compare the model-induced tree with the tree parsed by a benchmark parser, e.g., Berkeley Parser.
Specifically, we use the Berkeley parser to parse the same generated hypotheses by our model and treat the corresponding parsing results as golden parses.
Quantitative results (Figure \ref{tab:results_parser}) show that our model achieves an average F1 score of \textbf{94.6} 
, which demonstrates the generation process highly corresponds to the syntactic grammar and thus can be effectively interpreted.
Note that the score for WMT'14 En-De is lower (89.0), possibly due to the less accurate German parser for constructing the syntactic grammar, as discussed in Section~\ref{sec:grammar}.




\subsection{Controllability}
\label{sec:ana_control}

\begin{table}[t]
\centering
\small
\begin{tabular}{ccccc}
    \toprule
     \multirow{2}{*}{\textbf{Dataset}} &  \multicolumn{2}{c}{\textbf{BLEU} $\uparrow$}  & \multicolumn{2}{c}{$\mathbf{D}_{syn}^{ref}$ $\downarrow$} \\
     & \textbf{w/o} & \textbf{w} & \textbf{w/o} & \textbf{w} \\
     \midrule
     ParaNMT-small &  19.3 & 24.9(+5.6)   &  25.7 & 17.2(-8.5)\\
     NIST (ref-0) &  28.0  &  30.3(+2.3)   &  25.1 & 19.2(-5.9) \\
     NIST (ref-1) &   27.3 &  29.3(+2.0)   & 25.5  &  20.1(-5.4)\\
     NIST (ref-2) &   25.7 &  28.5(+2.8)   &  25.4 &  18.3(-7.1) \\
     NIST (ref-3) &  26.1  &  28.1(+2.0)   &  25.7 &  20.1(-5.6) \\
    WMT'16 Ro-En &  35.0 &  35.8(+0.8)  &  18.3 & 15.9(-2.4) \\
    WMT'14 De-En &  32.2 & 35.3(+3.1)   &  19.6 & 14.0(-5.6)\\
    WMT'14 En-De &  28.7 &  30.6(+1.9)  &  28.9 & 26.3(-2.6) \\
\bottomrule
\end{tabular}
        \caption{Controllable generation using golden syntax exemplars. NIST (ref-$i$) denotes the merged test sets with the $i$-th reference. A lower $\mathbf{D}_{syn}^{ref}$ denotes higher syntactic similarity with the reference.}
    \label{tab:results_control}
\end{table}

\paragraph{Control with Complete Syntax Template}
To leverage control signals from delexicalized syntax templates (e.g., ``(\texttt{S} (\texttt{NP}) (\texttt{VP} (\texttt{NP})))'' for the sequence ``I ate an apple.''), 
we introduce a reward $\gamma$ into Equation \ref{eq:weight_beam}:
\begin{equation}
     \delta_{ik+j} = \alpha 
 \delta_{\boldsymbol{s}_i} + (1-\alpha) \delta_{\boldsymbol{f}_{j;\boldsymbol{s}_i}} + \gamma.
\end{equation}
If the updated syntax context $\boldsymbol{s}_{ik+j}$  matches the corresponding template pattern at depth $d+1$, the $\gamma$ is a positive value otherwise 0.
For example, the syntax context ``<\texttt{NP}> <\texttt{VP}>'' in Figure \ref{fig:beam} matches the pattern ``((\texttt{NP})(\texttt{VP}))'' at depth 2.
Intuitively, the reward encourages the model to favor beam candidates that match the syntax template.
We set the reward value as 0.32 based on validation results (Appendix \ref{app:reward}).
The testset of ParaNMT-small is provided with human-annotated exemplars and we use it to control generation, with results shown in Table \ref{tab:para}.
More generally, golden templates can be derived by parsing the reference sentences for each dataset with a parser (e.g., the Berkeley Parser).
We present the results in Table \ref{tab:results_control}.
Guided by the reference syntax template, our model obtains consistent improvement in terms of hypothesis similarity with references, which is reflected by the decreased syntax edit distance to the references, i.e., $\mathbf{D}_{syn}^{ref}$.
For the multi-reference dataset NIST Zh-En, our model can generate translations of different styles which are prompted by alternative syntax templates from multiple references.


\paragraph{Control with Partial Syntax Template}
\label{para:human_control}
We further explore whether the model can handle fine-grained arbitrary controls.
Specifically, we ask three annotators to modify the intermediate syntax contexts output by the model, based on the source input.
100 instances are randomly selected from the NIST Zh-En test set and each annotator gives different modifications for each instance.
The modified contexts are fed to the model to predict the infilling texts.
We then ask the annotators to evaluate whether their controls (i.e., modifications) are safely responded to by the model.
We show some of the control examples in Appendix \ref{app:control_exp}.
The average control success rate is 81\%, which demonstrates the capability of our model to handle arbitrary fine-grained controls.


\subsection{Diversity}
\label{sec:ana_div}

\paragraph{Beam Diversity}
We expect the model to generate diverse hypotheses under beam search, while also maintaining generation quality.
To this end, we measure the model's beam diversity by computing two average scores: 
(1) the average of the mutual diversity scores of every two of the beam candidates, i.e., $\mathbf{D}_{lex}^{beam}$ and  $\mathbf{D}_{syn}^{beam}$; 
(2) the average generation quality of the beam candidates, measured by BLEU scores.
The results for paraphrase generation are shown in Table \ref{tab:beam_div}.
In terms of generation quality, our model generates consistently better beam candidates on average than the baseline model.
Besides, we can see that structural beam search can yield more diverse beam candidates, indicated by the higher mutual diversity (i.e., $\mathbf{D}_{lex}^{beam}$ and $\mathbf{D}_{syn}^{beam}$) among beam candidates.

\begin{table}[t]
\centering
\small
\begin{tabular}{cccc}
    \toprule
      \multicolumn{4}{c}{\textbf{ParaNMT-small}}\\
     \textbf{Model} & \textbf{avg BLEU/iBLEU}  &  $\mathbf{D}_{lex}^{beam}$$\uparrow$ &  $\mathbf{D}_{syn}^{beam}$$\uparrow$  \\
     \midrule
     Transformer &  15.0/1.6 &  12.6  &  11.2 \\
     Our Method &  \textbf{16.9/7.1} &  \textbf{15.0}   &  \textbf{12.6} \\

\bottomrule
\end{tabular}
    \caption{Beam diversity measured by the average generation quality and the average mutual diversity among the beam candidates.
    }
    \label{tab:beam_div}
\end{table}

\begin{figure}[t!]
\centering
\includegraphics[width=0.8\linewidth]{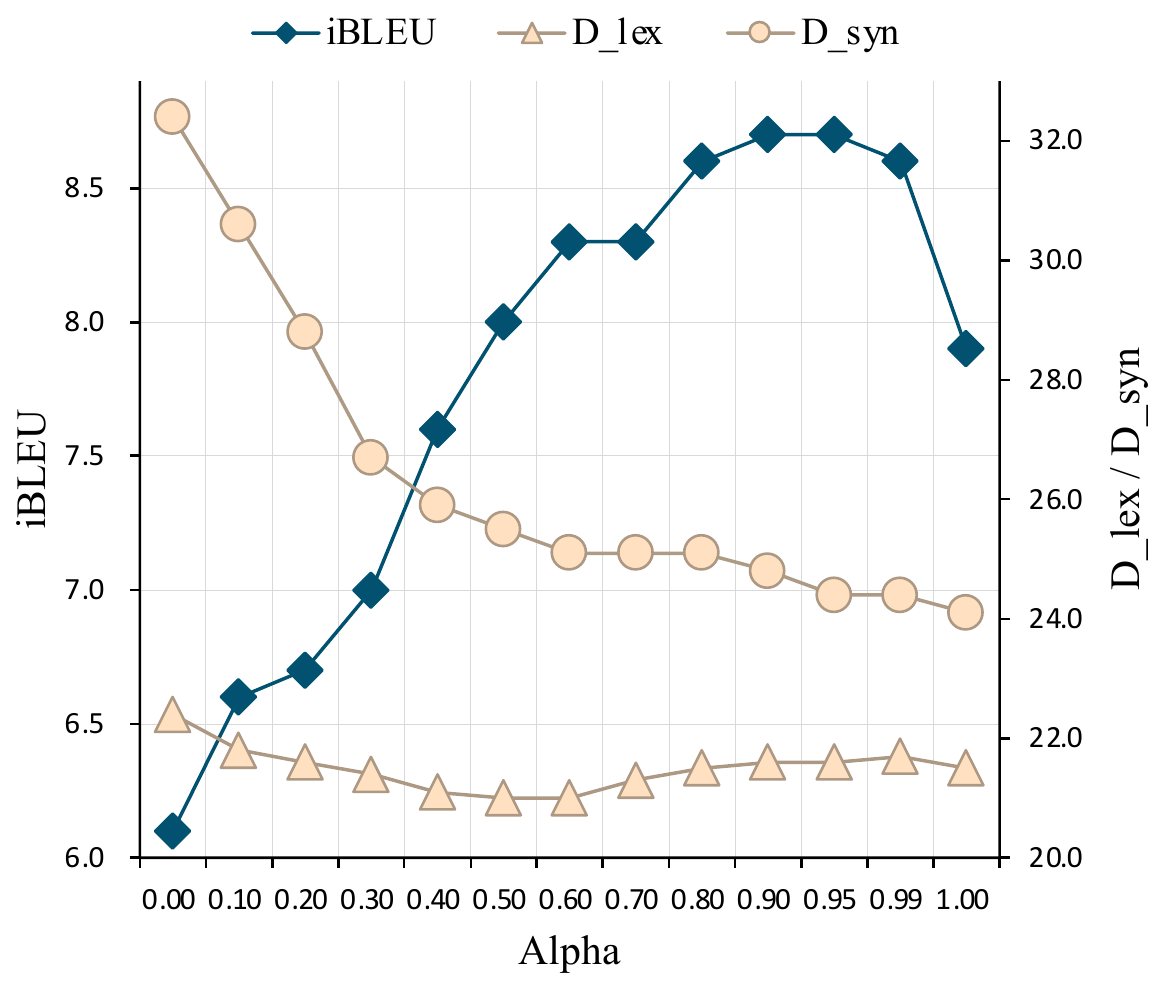}
    \caption{Effects of accumulation weights.}
    \label{fig:weight}
\end{figure}

\paragraph{Effects of Accumulation Weight}
A larger accumulation weight ($\alpha$ in Eq. \ref{eq:weight_beam}) indicates a larger weight on previous decisions when re-ranking the newly updated beam candidates.
As a result, early determined syntax structures are less likely to be surpassed throughout the whole structural beam search.
On the contrary, a smaller $\alpha$ encourages the model to explore promising candidates at higher levels, and can therefore find more diverse hypotheses.
We explore the effects of $\alpha$ with results shown in Figure \ref{fig:weight}.
As the weight grows smaller, the model generates sequences of better syntactic diversity, i.e., $\mathbf{D}_{syn}$.
However, an overly small weight deteriorates generation quality (iBLEU), which can be caused by the model's overconfidence in local predictions without considering the predictions of syntax contexts at lower levels.
Such deterioration is also seen for overly large weights (>0.95), due to limited exploration at higher levels.
\paragraph{Human Evaluation}
\label{para:human_eval}
\begin{table}[t]
\centering
\small
\begin{tabular}{cccc}
    \toprule
     \textbf{Model} & \textbf{Fidelity}& \textbf{Novelty} &\textbf{Diversity}  \\
     \midrule
     Transformer &  50.2\% & 29.6 \%   &  29.0\%  \\
     Our Method &  49.8\% & \textbf{70.4\%}   & \textbf{71.0\% } \\
\bottomrule
\end{tabular}
    \caption{Human evaluation on paraphrase generation.}
    \label{tab:para_human}
\end{table}
We further conduct a human evaluation to evaluate generation quality and diversity on paraphrase generation.
We ask three annotators to vote for one of the two candidates: hypotheses from the seq2seq baseline and our method.
The annotators are required to decide, which one is better by considering \textit{Fidelity}, \textit{Novelty}, and \textit{Diversity} (See Appendix \ref{app:para_human} for details). The results are shown in Table \ref{tab:para_human}. 
As can be seen from the table, our method achieves much better generation novelty and beam diversity compared with the baseline,
while maintaining semantic fidelity, which further validates the results of the automatic evaluation.




\section{Conclusion}
We proposed a syntax-guided generation paradigm, which leverages the strengths of Transformer and generates sequences by hierarchically expanding constituents in the lexicalized syntax contexts throughout the syntax tree.
The neural decoder was trained by maximizing the likelihood of the infilling texts for each constituent in the syntax contexts given the source sequence.
Moreover, we proposed the structural beam search to better explore the hypothesis space.
Empirical results demonstrated the advantage of generation quality over the seq2seq baseline, and also the effectiveness in terms of interpretability, controllability, and diversity.

Our method can be seen  as a step towards explicit modelling of psycholinguistic structures during neural text generation
, helping the model to have a degree of control over what it intends to generate, 
which can potentially address salient issues of current neural NLG,
such as hallucination \cite{hallucinationmt,hallucinationdialog} and ethical issues \cite{ethic1,ethic2,ethic3},
if semantics, pragmatics, and other factors are also integrated.

\section*{Limitations}
Despite the competitive performance, there are several limitations of this work:
(1) As discussed in Section~\ref{sec:grammar}, the generation performance relies on the parser performance, which is strong enough for English but still less satisfactory for other languages.
Dedicated methods need to be considered to compensate for the weak parser performance if we want to extend our method to more languages.
(2) In this work, we consider two NLG tasks with semantic equivalence to testify if the proposed method can convey the source semantics accurately by following the target syntactic grammar.
Other tasks such as summarization and dialogue generation can also be tested, where the semantics are not equivalent between the source and target.
(3) To train the neural decoder parallelly, we break down the source-target dataset into a triple set.
However, the global dependency of the syntax parse tree is not considered, which can deteriorate generation performance.
(4) Due to the recursive encoding of the syntax contexts, our model's inference speed is approximately half that of the seq2seq counterpart (Appendix \ref{app:gen_linear}).
(5) Future work should include experiments on large language models \cite{gpt3,gpt4,glm,llama,alpaca}. to further demonstrate the effectiveness of our method beyond pre-trained language models.

\section*{Ethics Statement}
We honor the ACL Code of Ethics. 
No private data or non-public information is used in this work.
For human annotation (Section~\ref{para:human_control} and Section~\ref{para:human_eval}), we recruited our annotators from the linguistics departments of local universities through public advertisement with a specified pay rate. 
All of our annotators are senior undergraduate students or graduate students in linguistic majors who took this annotation as a part-time job. We pay them 60 CNY an hour. The local minimum salary in the year 2022 is 25.3 CNY per hour for part-time jobs.
The annotation does not involve any personally sensitive information. The annotated is required to rank the system output and label factual information (i.e., syntactic annotation).

\section*{Acknowledgement}
We would like to thank all reviewers for their insightful comments and suggestions to help improve the paper. We thank Deng Cai and Xinting Huang for their insightful suggestions. 
This work is funded by the Ministry of Science and Technology of China (grant No. 2022YFE0204900).

\bibliography{acl}
\bibliographystyle{acl}
\clearpage
\appendix

\section{Algorithms}
\label{app:alg}

The scoring algorithm \ref{eq:score} can be rewritten with the source context $\mathbf{x}$ encoded into $\mathbf{h}_{src}$:
\begin{equation}
    \mathrm{score}(\mathbf{h}_{src},\boldsymbol{s},\boldsymbol{f})=\sum_{t=0}^{|\boldsymbol{f}|}logp_{\boldsymbol{\theta}}(f_t|\mathbf{h}_{src},\boldsymbol{s},f_{1:t-1})
    \label{eq:score2}
\end{equation}

The algorithm of \textbf{structural beam search} is demonstrated in Algorithm \ref{alg:beam}, which employs the standard beam search for autoregressive generation, depicted in Algorithm \ref{alg:std_beam}.
The termination function in Algorithm \ref{alg:beam} (i.e., $\mathrm{terminated}(\cdot)$) returns true if the there is no remaining constituent in the input sequence.

\algrenewcommand{\algorithmiccomment}[1]{\hskip1em$\rightarrow$ \footnotesize#1 \normalsize}
\begin{algorithm}[t]
\small
\textbf{Setup:} $k$: beam size \\
\hspace*{2.7em} $\alpha$: accumulation weight \\
\hspace*{2.7em} $d_{max}$: maximum tree depth \\
\hspace*{2.7em} $\textsc{Encoder}(\cdot)$: source context encoder \\
\hspace*{2.7em} $\mathrm{terminated(\cdot)}$: termination examination function  \\
\hspace*{2.7em} $\mathrm{expand(\cdot, \cdot)}$: constituent expansion function  \\
\hspace*{2.7em} $\mathrm{beam\_search(\cdot, \cdot)}$: standard beam search algorithm \\
\textbf{Input:} $\mathbf{x}$: source sequence
\begin{algorithmic}[1]
\State $d \gets 0$
\State $\mathbf{h}_{src} \gets \textsc{Encoder}(\mathbf{x})$
\State $B_0 \gets \{ (  0, \langle T \rangle ) \}$
\While { $d<d_{max}$ }
    \State $B \gets \emptyset$
    \For{$ ( \delta_{\boldsymbol{s}}, \boldsymbol{s} ) \in B_{d-1}$}    
    \If{$\mathrm{terminated(\boldsymbol{s})}$}
        \State $B.\mathrm{add}(( \delta_{\boldsymbol{s}}, \boldsymbol{s}  ))$
        \State \textbf{continue}
    \EndIf
    \State $\mathcal{F} \gets \mathrm{beam\_search(\boldsymbol{s},\mathbf{h}_{src})}$
    \For{$(\delta_{\boldsymbol{f}},\boldsymbol{f}) \in \mathcal{F}$}
        \State $\hat{\delta} \gets \alpha  \delta_{\boldsymbol{s}} + (1-\alpha) \delta_{\boldsymbol{f}}$ 
        \State $\hat{\boldsymbol{s}} \gets \mathrm{expand}(\boldsymbol{s},\boldsymbol{f})$
        \State $B.\mathrm{add}( ( \hat{\delta}, \hat{\boldsymbol{s}} ))$
    \EndFor
    \EndFor
    \State $B_d \gets B.\mathrm{top}(k)$ 
    \State $d \gets d + 1$
\EndWhile
\State \Return $B_{d_{max}}$
\end{algorithmic}
\caption{Structural beam search}
\label{alg:beam}
\end{algorithm}



\algrenewcommand{\algorithmiccomment}[1]{\hskip1em$\rightarrow$ \footnotesize#1 \normalsize}
\begin{algorithm}[tb]
\small
\textbf{Setup:} $k$: beam size \\
\hspace*{2.7em} $t_{max}$: maximum hypothesis length \\
\hspace*{2.7em} $\mathcal{V}$: target tokens set \\
\hspace*{2.7em} $\mathrm{score}(\cdot, \cdot, \cdot)$: scoring function (Eq. \ref{eq:score2})\\
\textbf{Input:} $\boldsymbol{s}$: syntax context \\
\hspace*{2.7em} $\mathbf{h}_{src}$: source context representations
\begin{algorithmic}[1]
\State $t \gets 0$
\State $B_0 \gets \{ (  0, \langle bos \rangle ) \}$
\While { $t<t_{max}$ }
    \State $B \gets \emptyset$
    \For{$  (\delta, \boldsymbol{f}) \in B_{t-1} $}    
    \If{$\boldsymbol{f}.last() = \langle eos \rangle $}
        \State $B.\mathrm{add}(( \delta, \boldsymbol{f}  ))$
        \State \textbf{continue}
    \EndIf
    \For{$f \in \mathcal{V}$}
        \State $\delta \gets \mathrm{score}(\mathbf{h}_{src},\boldsymbol{s},\boldsymbol{f} \circ f)$ 
        \State $B.\mathrm{add}( ( \delta, \boldsymbol{f} ))$
    \EndFor
    \EndFor
    \State $B_t \gets B.\mathrm{top}(k)$ 
    \State $t \gets t + 1$
\EndWhile
\State \Return $B_{t_{max}}$
\end{algorithmic}
\caption{Beam search}
\label{alg:std_beam}
\end{algorithm}

\begin{table*}[t!]
\small
    \scriptsize{\centering
        \begin{tabular}{cccccccccc}
            \toprule
              \textbf{Model} & \textbf{BLEU}$\uparrow$  / \textbf{self-BLEU}$\downarrow$  / \textbf{iBLEU}$\uparrow$ & \textbf{METEOR}$\uparrow$ & \textbf{ROUGE-1/2/L}$\uparrow$ & $\mathbf{D}_{lex}$$\uparrow$  &  $\mathbf{D}_{syn}$$\uparrow$ \\
            \midrule
                BART Seq2seq (beam 1) & 15.8 / 26.9 / 3.0 & 27.3 & 50.1 / 23.1 / 50.0 & 19.5 &  23.8 \\
                BART + Our Method (beam 1) & 18.3 / 15.5 / 8.2 & 31.0 & 52.1 / 26.7 / 52.1 & 21.1 & 24.0 \\
                BART Seq2seq (beam 5) & 17.9 / 27.0 / 4.4 & 28.4 & 51.4 / 24.8 / 51.5 & 19.6 & 24.4\\
                BART + Our Method (beam 5) & \textbf{19.0} / \textbf{15.1} / \textbf{8.8} & \textbf{31.3} & \textbf{52.3} / \textbf{27.0} / \textbf{52.5} & \textbf{21.3} & \textbf{24.7} \\
            \midrule  
        \end{tabular}
        \caption{Experimental results on paraphrase generation (ParaNMT-small) based on BART.}
    }

\label{tab:bart-full}
\end{table*}

\section{Experiment Details}
\label{app:setup}
For NIST Zh-En, we use parts of the bitext provided within NIST’12 OpenMT\footnote{LDC2005T06, LDC2004T07, LDC2003E07, LDC2000T46, LDC2000T47, LDC2000T50, LDC2003E14, LDC2005T10, LDC2002E18, LDC2007T09, LDC2004T08} and the final train set consists of about 1.8M sentence pairs.
We apply BPE \cite{BPE} on all datasets: the number of BPE operations is 6K for ParaNMT-small, and 40K for the other datasets.
We implement our model using Fairseq \cite{fairseq}.
We train the model using Adam \cite{adam} optimizer. 
The learning rate increases to $7\cdot10^{-4}$ in the first 10K steps and then anneals exponentially. 
We set the weight decay as 0.01 and label smoothing as 0.1.
The dropout is 0.3 for ParaNMT-small, and 0.1 for the other datasets.
The batch size is 64K tokens for ParaNMT-small, 256K for WMT'16 Ro-En and NIST Zh-En, and 512K for WMT'14 De$\leftrightarrow$En.
All models are trained for a maximum update of 300K steps unless early stopped.
We train the model using 4 V100s and increase gradient accumulation steps for large batch sizes. 
We choose the 5 best checkpoints based on validation sets and average them for inference.
We set the beam width as 5 for beam search.
For machine translation, the teacher models for knowledge distillation are Transformer\_Base for NIST Zh-En and WMT'16 Ro-en, and Transformer\_Big for WMT'14 De$\leftrightarrow$En.

\section{Model Architecture}
\label{app:arch}
We conduct experiments to compare different model architectures to incorporate syntax context on the WMT'16 Ro-En validation set. We consider the following settings:
\begin{itemize}
    \item \textit{Concat}: concatenate the syntax context with the source sequence, with the vanilla Transformer unmodified.
    \item \textit{Extra-attention}: reuse the source encoder for encoding syntax context and insert an extra attention layer, i.e., the syntax context attention, into each decoder layer.
    \item \textit{Extra-encoder}: introduce an additional encoder for encoding syntax context and also uses the syntax context attention.
\end{itemize}
Empirical results are shown in Table \ref{tab:arch}.
Based on validation results, we adopt the \textit{Extra-encoder} model in all experiments except for training on BART (Table \ref{tab:bart}), where we adopt the \textit{Concat} model.

\begin{table}[t]
\centering
\small
\begin{tabular}{cccc}
    \toprule
     \textbf{Architecture} & \textbf{\# params} &\textbf{BLEU}  & \textbf{Speed}  \\
     \midrule
     Concat & 64.2M & 34.5  & 1.0x\\
     Extra-attention & 70.5M & 34.7 & 0.9x\\
     Extra-encoder & 64.2M & \textbf{35.3} & \textbf{1.1x}\\
\bottomrule
\end{tabular}
    \caption{Model architectures for encoding previous syntax contexts.}
    \label{tab:arch}
\end{table}

\section{Experiments on PLM}
\label{app:bart}
In this section, we introduce our experiment settings of PLM. Following previous work~\cite{sun-etal-2021-aesop}, we use BART-base~\cite{bart} as our base model. All models are finetuned for 10 epochs with a batch size of 64k tokens. The learning rate is 3e-5 and the linear decay schedule, as recommended in BART's official repository\footnote{\url{https://github.com/facebookresearch/fairseq/tree/main/examples/bart}}. 

We use the \textit{Concat} (Appendix \ref{app:arch}) model architecture for extending our method to BART.
The source text and the syntax context are concatenated with a special token ``<\texttt{sep}>'', e.g., ``I ate an apple . <\texttt{sep}> <\texttt{NP}> <\texttt{VP}> .''.
To effectively employ our method with BART, whose inputs are tokenized sequences byte-level, as same as \citet{gpt2}, we make several modifications. In the pre-processing, we make sure our special tokens (e.g., <\texttt{sep}>, <\texttt{c}>, <\texttt{NP}>, <\texttt{VP}>) are not split and add extra byte-level spaces before and after the special token. Thanks to the unused tokens in BART embeddings, we do not need to modify the embedding matrix. 
Instead, we assign our special tokens to unused token indexes. 
Finally, in the inference stage, we find the constituency expansion causes a discrepancy between inputs of train and test. Thus, we first detokenize each layer's outputs and then tokenize them back with the same procedure in the preprocessing to avoid such a gap.


\begin{figure}[t!]
\small
\setlength{\belowcaptionskip}{-0.5cm}
\centering
\includegraphics[width=0.7\linewidth]{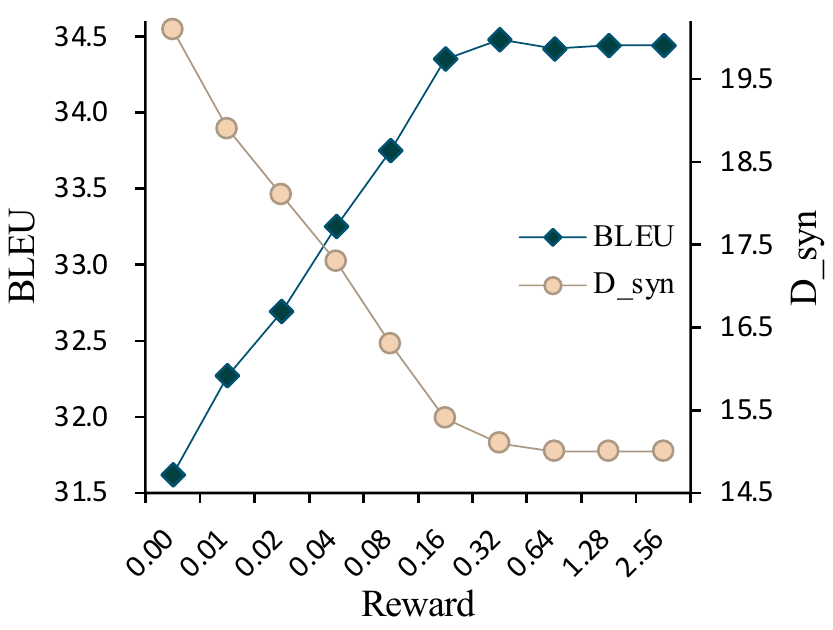}
    \caption{Effects of reward ratio on the WMT14'De-En validation set.}
    \label{fig:reward_ratio}
\end{figure}
\begin{figure*}[t!]
\small
\centering
\includegraphics[width=0.99\linewidth]{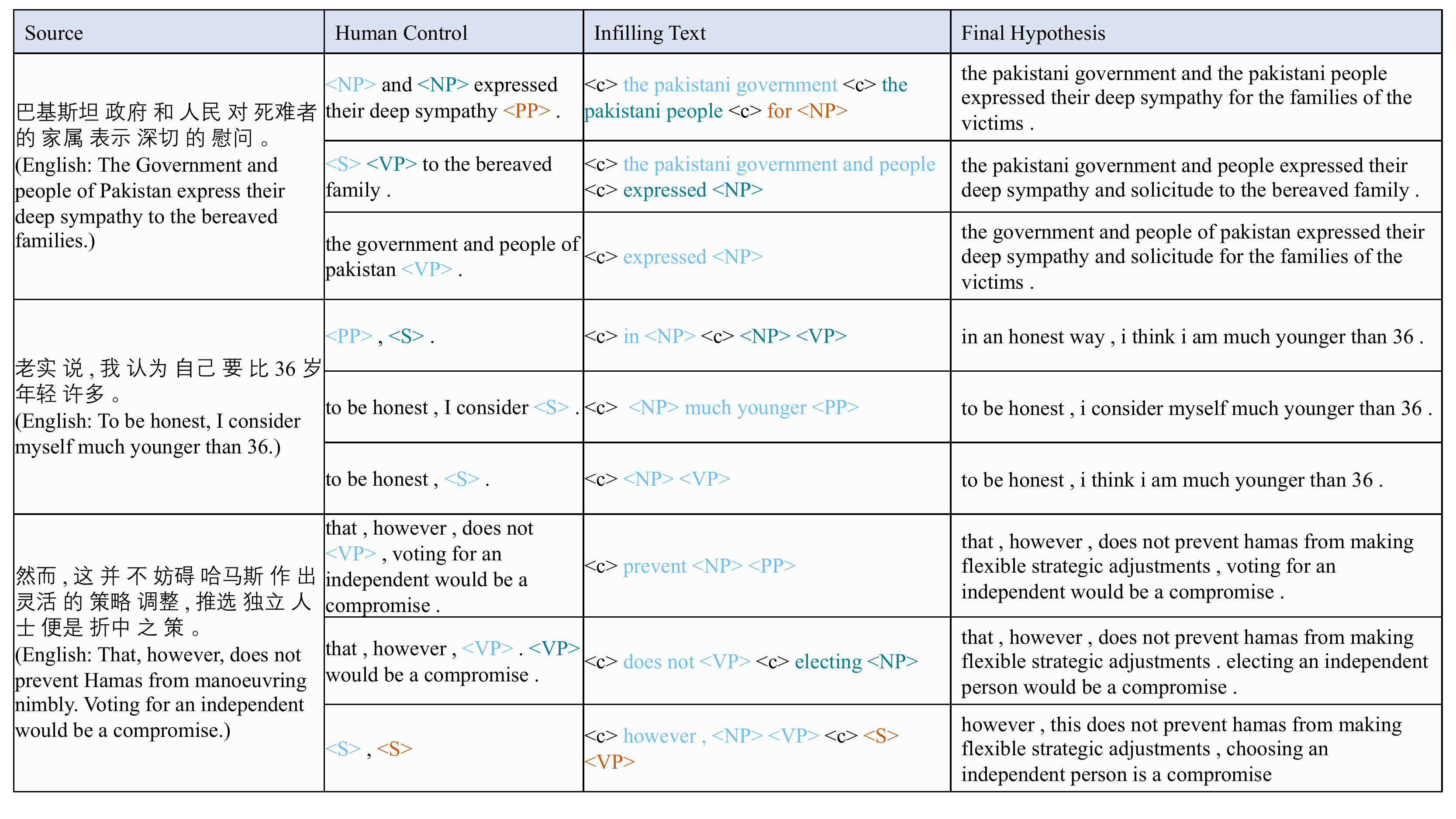}
    \caption{Samples cases for fine-grained manual controls: the 4 columns denote the source Chinese sentence, the human-annotated control, the model's predicted infilling texts, and the final English translation.}
    \label{fig:control_case}
\end{figure*}

\section{Generating Linearized Trees Directly}
\label{app:gen_linear}
A baseline method to induce grammar simultaneously during generation is generating linearized parse trees directly, i.e., training a seq2seq model which takes in source sequences and outputs linearized parse trees.
We compare it with our method on WMT'16 Ro-En.
Specifically, the BLEU score for WMT'16 Ro-En is only \textbf{27.6} compared to the seq2seq baseline (\textbf{34.1}) and our method (\textbf{34.9}). 
This can be because the additional parentheses and constituency tags in linearized trees may deteriorate sequence coherence, making learning more difficult. 
Our method, on the other hand, breaks down syntax trees into level pieces to create a better learning curriculum.
Furthermore, Generating linearized parse trees is much slower than the seq2seq counterpart, since the average sequence length of linearized tree sequences is longer (152.3 vs 28.4). 
As a result, the average speed for generating linearized parse trees is only 0.8 sentences/s compared to 3.6 sentences/s for the seq2seq baseline.
Our method achieves an inference speed of 1.7 sentences/s under the same computing condition (V100 GPU). 
Additionally, generating a linearized parse tree is not easily interpretable or controllable, due to the black-box nature of the sequence-to-sequence paradigm.

\section{Effects of Control Reward}
\label{app:reward}
The magnitude of the reward $\gamma$ determines how much priority is given to beam candidates that match the syntax exemplar. We experiment with different reward values to give a quantitative demonstration, shown in Figure \ref{fig:reward_ratio}.
It can be seen that the control effectiveness grows with the increase of the reward value until 0.64, which suggests that all possible matched beam candidates are re-ranked to the top in the search space.

\section{Control with Partial Syntax Template}
\label{app:control_exp}

We present 3 sample cases to demonstrate fine-grained controls over the generation process, shown in Figure~\ref{fig:control_case}.
Each Chinese source sentence is paired with 3 manual controls from three annotators.
The model takes in the annotated syntax context and proceeds to obtain the respective translations.

\section{Human Evaluation for Paraphrase Generation}
\label{app:para_human}

We ask three annotators to conduct side-by-side human evaluations and report averaged results of their annotations.
For each instance, the annotators vote for one of the two outputs by the baseline and our model.
The outputs contain top-5 beam candidates under beam search.
The annotators are asked to evaluate both the best candidate and the beam results as a whole,
based on the following three aspects:

\begin{itemize}
    \item Fidelity: Whether the best candidate is semantics-equivalent with the input.
    \item Novelty: Whether the best candidate modifies the input sentence structure.
    \item Diversity: Whether the generated five candidates are different from each other given the input.
\end{itemize}

\end{document}